\documentclass{article}

\PassOptionsToPackage{numbers, compress}{natbib}

\usepackage[final]{neurips_2023}
\usepackage{graphicx}
\usepackage{amsmath}
\usepackage{amssymb}
\usepackage{algorithm}
\usepackage{algpseudocode}
\usepackage{multirow}
\usepackage{caption}
\usepackage{subcaption}

\usepackage[utf8]{inputenc} 
\usepackage[T1]{fontenc}    
\usepackage{hyperref}       
\usepackage{url}            
\usepackage{booktabs}       
\usepackage{amsfonts}       
\usepackage{nicefrac}       
\usepackage{microtype}      
\usepackage{xcolor}         

\title{Improving Deep Generative Models on Many-To-One Image-to-Image Translation}

\author{%
  Sagar Saxena\\
  Department of Computer Science\\
  University of Maryland\\
  College Park, MD 20742 \\
  \texttt{ssaxena1@umd.edu} \\
  \AND
  Mohammad Nayeem Teli\\
  Department of Computer Science\\
  University of Maryland\\
  College Park, MD 20742 \\
  \texttt{nayeem@cs.umd.edu} \\
}

\begin{document}

\maketitle

\begin{abstract}
  Deep generative models have been applied to multiple applications in image-to-image translation. Generative Adversarial Networks and Diffusion Models have presented impressive results, setting new state-of-the-art results on these tasks. Most methods have symmetric setups across the different domains in a dataset. These methods assume that all domains have either multiple modalities or only one modality. However, there are many datasets that have a many-to-one relationship between two domains. In this work, we first introduce a Colorized MNIST dataset and a Color-Recall score that can provide a simple benchmark for evaluating models on many-to-one translation. We then introduce a new asymmetric framework to improve existing deep generative models on many-to-one image-to-image translation. We apply this framework to StarGAN V2 and show that in both unsupervised and semi-supervised settings, the performance of this new model improves on many-to-one image-to-image translation.
\end{abstract}

\section{Introduction}

Many recent works in image-to-image translation have used deep generative models to learn state-of-the-art mappings between multiple domains. These models have been used for a wide array of tasks: colorization, style transfer, semantic image synthesis, image completion, super resolution, inpainting, and many more \cite{comp_analysis_survey, multi-modal_survey, deep-gen_survey, gen-adv-survey, adv-learning-survey, image-synthesis-survey}. Most approaches propose architectures that model this mapping as bijective \cite{cyclegan, discogan, dualgan, unit, pix2pix, cogan, stargan}, where each image from one domain can only map to one image in another domain, or many-to-many \cite{munit, drit, starganv2, palette}, where each image from one domain can map to many in another domain. The ability to synthesize multiple images from one image is usually regarded as multi-modal image-to-image translation. 

Many of the datasets that these methods are trained on, however, do not have bijective or many-to-many relationships between domains. In tasks such as image colorization \cite{palette}, semantic segmentation \cite{cityscapes, synthia, facades, multi-ped, coco}, depth estimation \cite{synthia, nyu-depth, rgbd}, heat mapping \cite{multi-ped}, edge image synthesis \cite{pix2pix, cyclegan}, etc., a model that learns either a bijective or many-to-many mapping between domains would not accurately model the relationship between domains. Instead of learning either uni-modal or multi-modal synthesis for all domains, a good model would learn uni-modal synthesis for domains that only have one representation for each input image and multi-modal synthesis for domains that have multiple image representations for each input image. 

\begin{figure}[!h]
  \centering
   \includegraphics[width=0.5\linewidth]{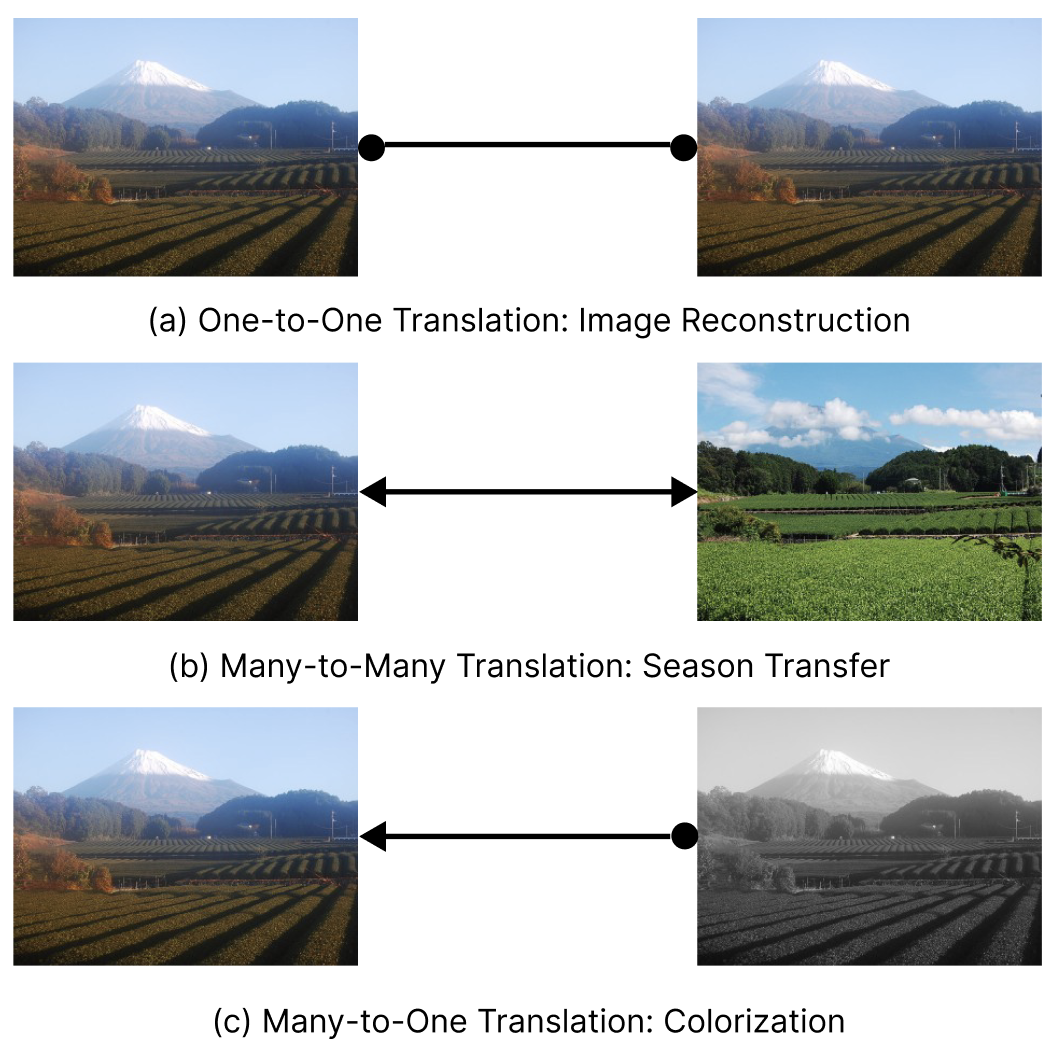}

   \caption{Different relationships between domains on \cite{trans-attr}}
   \label{fig:modalities}
\end{figure}

To address this problem, we propose that if the domains have different types of modalities, the architecture of a model and its loss functions should also reflect this. In this paper, we discuss the development of deep generative image-to-image translation methods and why they may not be sufficient for datasets with many-to-one relationships between domains in Section~\ref{sec:related-work}. Our contributions include

\begin{itemize}
    \item introducing a new framework that optimizes for many-to-one image translation and applying it to StarGAN V2 \cite{starganv2} in Section~\ref{sec:opt},
    \item providing a Colorized MNIST dataset and introducing a Color Recall metric that is more interpretable than existing metrics for ranking different architectures on many-to-one image translation tasks in Section~\ref{sec:color-mnist},
    \item demonstrating, with experimental results, that the Colorized MNIST dataset can provide interpretable insights on a model's ability to generate diverse images in Section~\ref{sec:color-mnist-results},
    \item demonstrating, with experimental results on the Colorized MNIST and ADE20K \cite{ade20k1, ade20k2} datasets, that our method is able to either acheive better performance or a better trade-off when generating images in multiple domains with different modalities in Section~\ref{sec:results}
\end{itemize}


\section{Related Work}
\label{sec:related-work}

\subsection{Generative Adversarial Networks}
Image-to-image translation with deep generative models has focused on Generative Adversarial Networks \cite{gan} (GANs). GANs generate realistic images by training a generator that generates images and an adversarial discriminator that discriminates whether images are real or fake. Initial work used paired data \cite{pix2pix, cogan} to create a bijective mapping between two domains by using a supervised loss function to ensure that each $x \in D_x$ mapped to a known $y \in D_y$. 

To address the lack of paired data, the initial works were expanded to use unpaired data to learn bijective mappings \cite{cyclegan, discogan, dualgan, unit}. Two sets of generators and discriminators are trained in \cite{cyclegan, discogan, dualgan}, with a cycle consistency loss to learn that each $x \in D_x$ maps to some $y \in D_y$ and back to the known $x \in D_x$. In \cite{unit, stargan}, there exists a shared latent space for all domains: for each unknown pair $(x, y)$ there exists a shared $z$ where $z$ maps to $x \in D_x$ and $y \in D_y$. 

More recent work uses unpaired data to learn many-to-many mappings. In \cite{munit, drit, starganv2}, an additional style encoder is used to extract the styles from each domain and generators are modified to use both the latent vector $z$ and a style $s$. This style space can be either sampled or the style of a guide image can be extracted. By adding these style encoders, GANs were able to generate diverse outputs for a given input image across all domains. In \cite{starganv2}, all weights are shared across domains and there is only one generator and discriminator for any number of domains. This increases scalability while reducing training time and model complexity.

Only a few works have attempted to leverage the many-to-one relationship between domains in a dataset \cite{spade, oasis}. In \cite{oasis}, the discriminator was modified to produce an output map that labeled each pixel as either its class (for a semantic segmentation task) or "fake". While this approach produced state-of-the-art results on challenging semantic image synthesis tasks, it required paired inputs to train.

\subsection{Diffusion Models}

Recent work in diffusion models \cite{denoising-diffusion} has shown that diffusion models can outperform GANs in image synthesis \cite{diffusion-beats-gans}. The diffusion process gradually adds controlled noise to an image $x \in D_x$ to generate a latent $z \in \mathcal{N}(0, I)$. Diffusion models generate realistic images by training a parametrized Markov chain to reverse this diffusion process. Several approaches have emerged recently that take advantage of conditional diffusion models~\cite{chenICLR2021, Saharia} on image-to-image translation tasks. 

In \cite{palette}, the authors trained a diffusion model to create images $y \in D_y$ in the target domain and used the corresponding pair $x \in D_x$ from the source domain as an additional input during the diffusion process. Similar to initial works in GANs \cite{pix2pix, cogan}, this model required paired examples and a supervised training paradigm. 

Similar to \cite{unit, stargan}, a few approaches learn a shared latent space across domains \cite{unit-ddpm, dual-diffusion} to allow for unsupervised image-to-image translation. In \cite{dual-diffusion}, two diffusion models are used for each domain - one for mapping to a shared latent space and one for mapping from the shared latent space to the target domain.

Similar to \cite{munit, starganv2}, \cite{diffusion-style} creates a disentangled latent space for content and style representations. \cite{diffusion-style} uses a complex approach that leverages vision transformers and contrastive and semantic style loss terms to generate high quality images. This indicates that, similar to GANs, new work in Diffusion Models will shift towards generating images from separate content and style latent spaces.

\section{Optimizing For Many-To-One Image-To-Image Translation}
\label{sec:opt}

The intuition for optimizing for many-to-one image-to-image translation is simple: if a dataset consists of domains that are asymmetric, model architecture should also be asymmetric. In both GANs and Diffusion Models, a few key loss terms and architectures are used to generate diverse and high fidelity images in all domains:

\begin{itemize}
    \item shared weights with conditional inputs allow learning shared representations across domains,
    \item separating style and content spaces allows "diversity" to be embedded in the style space,
    \item crafted loss terms can help introduce higher diversity and desirable reconstructions.
\end{itemize}

We propose a few simple modifications when working with a many-to-one task:

\begin{itemize}
    \item two domain-specific layers (one for decoding, one for encoding) should be used to map the original channel space to a shared channel space,
    \item for a uni-modal domain, only the content space should be used to generate the image,
    \item diversity should only be encouraged for multi-modal domains,
    \item optionally, when provided paired samples, a supervised loss should be assigned to the uni-modal domain.
\end{itemize} 

\subsection{Applying optimizations on StarGAN V2}

StarGAN V2 is one of the best performing approaches for GAN-based image-to-image translation. In this section, we present how the modifications outlined above can be applied to StarGAN V2. The base architecture for StarGAN V2 is slightly modified to use a Weight Demodulation layer instead of AdaIN \cite{adain} following \cite{stylegan2}.

StarGAN V2 consists of 4 modules: a style encoder $E$ that creates a style vector $s$ from an image $x$, a mapping network $F$ that creates a style vector $s$ from a latent vector $z \in \mathcal{N}(0,1)$, a generator $G$ that translates images given an input $x$ and style $s$, and a discriminator $D$ that outputs whether an image is real or fake given an image $\hat x$ and domain $d$.

First, we add four channel mapping components $C_{A \rightarrow S}, C_{S \rightarrow A}, C_{B \rightarrow S}, C_{S \rightarrow B}$ that encode and decode from the original image channel space (in domains $A$ and $B$) to a shared channel space ($S$). These are represented as trainable 1x1 convolutional layers. These networks are accompanied by simple cycle consistency loss terms:

\begin{equation}
    L_{ch\_cyc} = L1(C_{S \rightarrow A}(C_{A \rightarrow S}(x_A)), x_A) + L1(C_{S \rightarrow B}(C_{B \rightarrow S}(x_B)), x_B)
\end{equation}

Ideally, we want to only use the content space to create images in the uni-modal domain (for simplicity let us assume that this is domain $B$) while maintaining the weight sharing between generators for both domains. In each weight demodulation component, the style vector $s$ is inputted to a dense layer. The outputs of this dense layer scale the weights of the convolutional layer. Instead of directly modifying the components for the uni-modal domain, we can just use $s = 0$ for any uni-modal domain. This will have the effect of scaling the convolutional weights by the bias term of the dense layer (regardless of what $E$ or $F$ produces) and creating a fixed network for the unimodal domain. It will also zero any gradients that update the weights of the dense layer.

The original StarGAN V2 consists of a diversity synthesis loss term:

\begin{equation}
    L_{ds} = - L1(G(x, s_1), G(x, s_2)), s_1 \neq s_2
\end{equation}

This loss term ensures that images that are generated with different styles are diverse. This, however, penalizes a model for correctly generating consistent images in a uni-modal domain. To address this, the diversity synthesis loss term can be rewritten as follows:

\begin{equation}
    L_{ds} = - L1(G(x_B, s_1), G(x_B, s_2)), s_1 \neq s_2 \neq 0 
\end{equation}

This ensures that we only penalize the model for diversity in the multi-modal domain. Throughout this paper, we will abbreviate this new model with the above modifications as HMU. 

Given that we have pairs of images $(x_A, x_B)$, we can also add the following supervised loss term:

\begin{equation}
    L_{sup} = L1(G(x_A, 0), x_B)
\end{equation}

Throughout this paper, we will abbreviate the HMU model with this additional loss term as HMS.

\section{Colorized MNIST Dataset}
\label{sec:color-mnist}

In this section, we introduce a Colorized MNIST dataset as a simple benchmark dataset for evaluating many-to-one image-to-image translation problems. 

\begin{figure}[!h]
  \centering
   \includegraphics[width=0.4\linewidth]{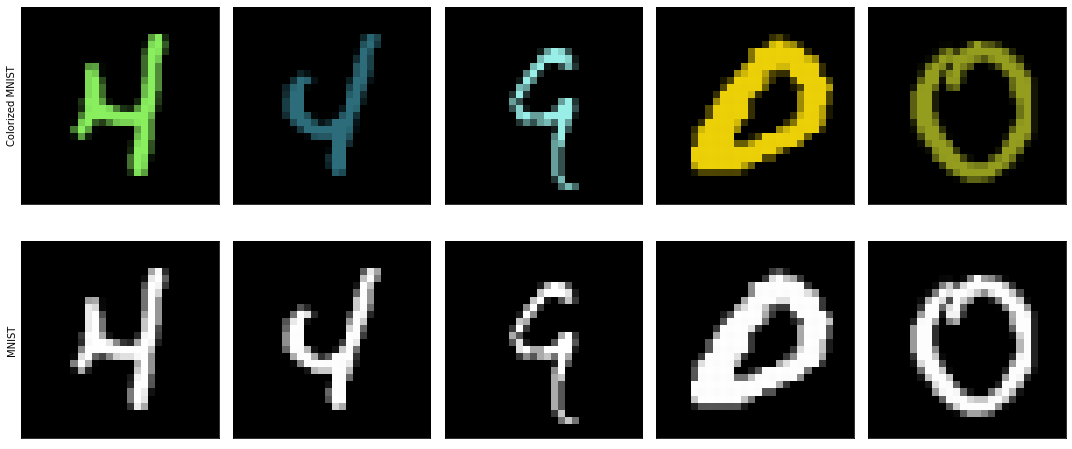}

   \caption{5 paired samples from the Colorized MNIST dataset.}
   \label{fig:color-mnist-samples}
\end{figure}

The MNIST \cite{mnist} dataset consists of 60k training images and 10k test images of 10 digits. For each MNIST image, a random color is selected to generate a 3-channel colorized MNIST image. 

\begin{equation}
  x_{a}^{(i)} = c_{i} * x_{b}^{(i)}, 
  x_{b}^{(i)} \in \textrm{MNIST}, 
  c_{i} \in \mathbf{R}^3
  \label{eq:color-mnist}
\end{equation}

Models are trained to map between standard MNIST images and colorized MNIST images (black/white $\leftrightarrow$ color). Unlike other colorization tasks, this task is unique in that it can be composed of two independent tasks: first, regenerate the input image, then, sample a random color and paint the regenerated image. The color of the output image is independent of the input image.

This has the effect of cleanly dividing the latent space. The latent space for content representations should only encode the information needed to reproduce the MNIST image. The latent space for style representations should only encode the color space.

A good model should be able to generate MNIST images that have been painted with colors that are uniformly sampled. This color can be extracted by finding the pixel with the maximum intensity in a given image. We can measure how good a model is at sampling colors uniformly by creating an n-binned Color Recall score for each color channel. 

\begin{algorithm}
\caption{Color Recall Score}
\label{alg:color-recall}
\begin{algorithmic}
\Require $n > 0$ \Comment{Number of bins}
\Require $c \in$ [Red, Green, Blue] \Comment{Color channel}
\Require $D$ \Comment{Dataset of real images}
\Require $\hat{D}$ \Comment{Dataset of generated images}
\Function{bin}{p} \Comment{The bin of a single pixel value}
  \State \textbf{return} $\lfloor\frac{p*n}{n}\rfloor$
\EndFunction

\Function{freq}{D} \Comment{Bin frequency in dataset $D$}
  \State $F_c \gets [0]$
  \ForAll{$x \in D$}
      \State $p_c \gets max(x_c)$
      \State $b \gets \mathrm{bin}(p_c)$
      \State $F_c[b] \gets F_c[b] + 1$ \Comment
  \EndFor
\EndFunction

\State $F_c \gets \mathrm{freq}(D)$
\State $\hat{F_c} \gets \mathrm{freq}(\hat{D})$
\State $s \gets \frac{1}{n}\sum_{b=0}^{n}{min(\frac{\hat{F_c[b]}}{F_c[b]}, 1)}$ \Comment{Average recall for each bin}
\State \textbf{return} s

\end{algorithmic}
\end{algorithm}

Another, simpler metric of interest is the number of unique colors that are in a generated dataset. Rather than having each pixel $p \in \mathbb{R}^3$, each pixel can be restricted to $p \in \mathbb{Z}^3 \cap [0, 255]$ (or 256 bins with a bin width of 1).

\begin{algorithm}
\caption{Unique Color Count}
\label{alg:unique-color-count}
\begin{algorithmic}
\Require $D$ \Comment{Dataset of images}

\State $S \gets \emptyset$

\ForAll{$x \in D$}
  \State $p_{rgb} \gets max(x)$ \Comment{pixel with maximum intensity}
  \State $S \gets S \cup {p_{rgb}}$ 
\EndFor

\State \textbf{return} $\lvert S \rvert$

\end{algorithmic}
\end{algorithm}

Both of these metrics are easier to interpret than FID \cite{fid}, Precision \cite{precision_recall_distributions}, and Recall \cite{precision_recall_distributions}, which rely on generating embeddings using a pre-trained neural network for real and generated images. Although this metric is tailored for the Colorized MNIST dataset, the goal of this section is not to present a new metric that works well on all datasets. Instead, this metric can be used in combination of the Colorized MNIST dataset to provide a more interpretable ranking and comparison of different models.

\section{Experiments and Results}
\label{sec:results}

\subsection{Datasets}

We show results on 2 datasets: the Colorized MNIST dataset detailed in Section~\ref{sec:color-mnist} and the more challenging ADE20K dataset \cite{ade20k1, ade20k2}. The ADE20K dataset contains 20k training and 2k test samples of RGB images of scenes and semantic segmentation maps of 150 different objects. 

\begin{figure}
    \centering
    \includegraphics[width=0.4\linewidth]{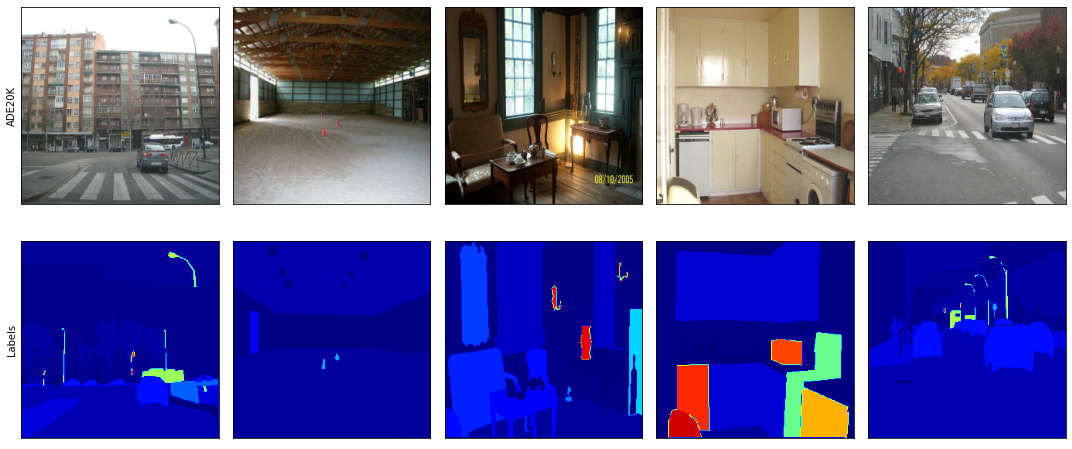}
    \caption{5 samples from the ADE20K training set.}
    \label{fig:my_label}
\end{figure}

\begin{figure}[!h]
    \centering
    \includegraphics[width=0.4\linewidth]{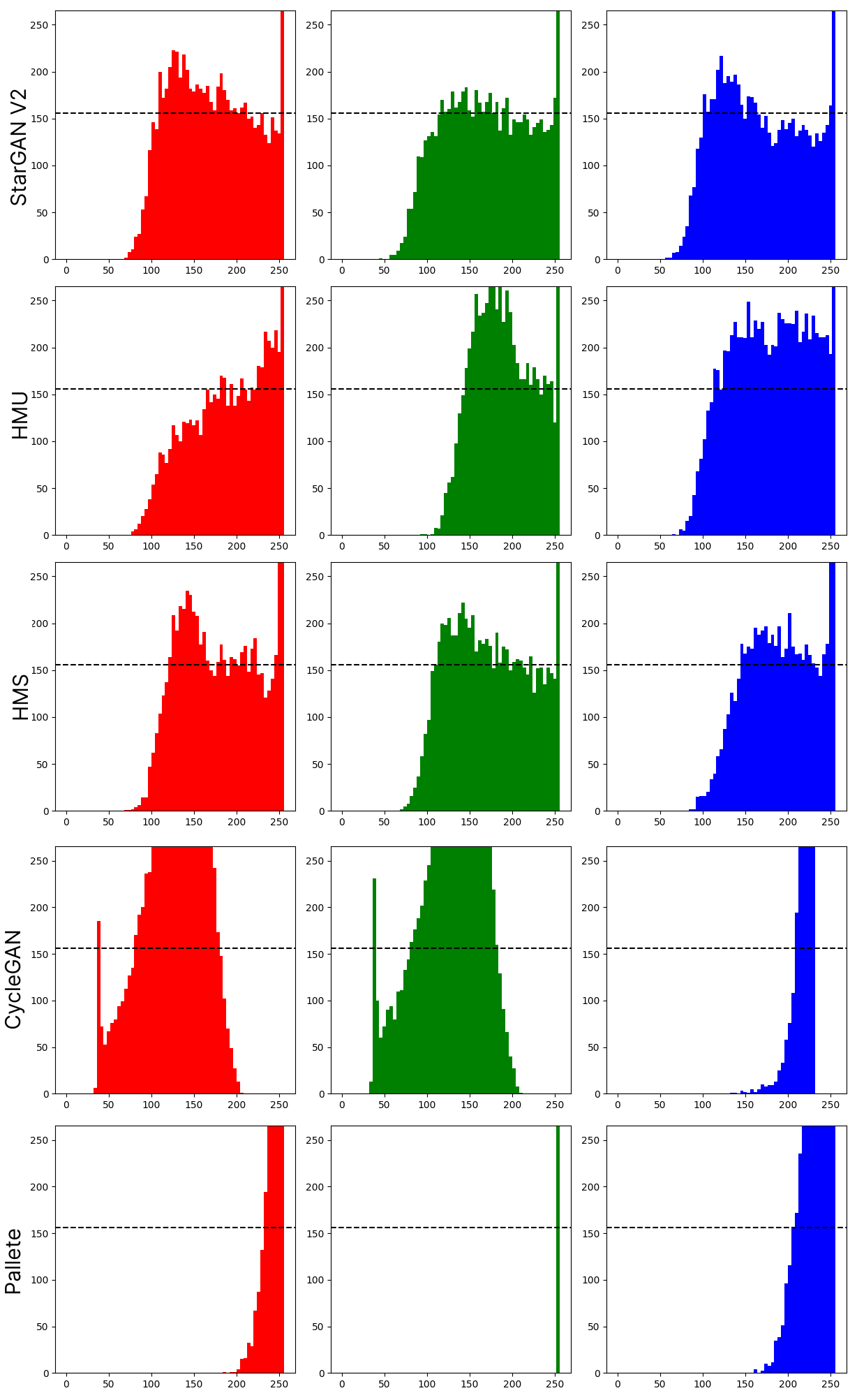}
    \caption{Red, Green, and Blue, Color Recall Histograms. The black dashed line represents the real uniform distribution of color values.}
    \label{fig:color-recall-dist}
\end{figure}

\subsection{Metrics}

Consistent with prior approaches, each model was evaluated on Fréchet Inception Distance \cite{fid} and Precision and Recall \cite{precision_recall_distributions} with embeddings generated using the InceptionV3 \cite{inception-v3} model. Precision and Recall curves are represented as a pair of two metrics, the F-8 and F-1/8 scores, as described in \cite{precision_recall_distributions}. For the Colorized MNIST dataset, we include metrics on the Color Recall score and Mean Squared Error on the MNIST domain. For the ADE20K dataset, classification accuracy has also been reported. All metrics are reported on the test sets.

\subsection{Training Details}

All models were trained on 4 NVIDIA RTX A6000 GPUs on 100 thousand samples of image pairs (200k images total). Data samples were scaled between -1 and 1 during training. StarGAN V2 and all of its modifications were reimplemented to run with Python 3.9.6, TensorFlow 2.8, CUDA 11.3.1, and cuDNN 8.2.1. All loss terms were weighted equally (with $\lambda = 1$). No additional modifications to the hyperparameters from the original StarGAN V2 paper (other than ones detailed in Section~\ref{sec:opt}).

\subsection{Colorized MNIST Results}
\label{sec:color-mnist-results}

\begin{figure}[]
    \centering
    \begin{subfigure}[]{0.4\textwidth}
        \centering        \includegraphics[width=\textwidth]{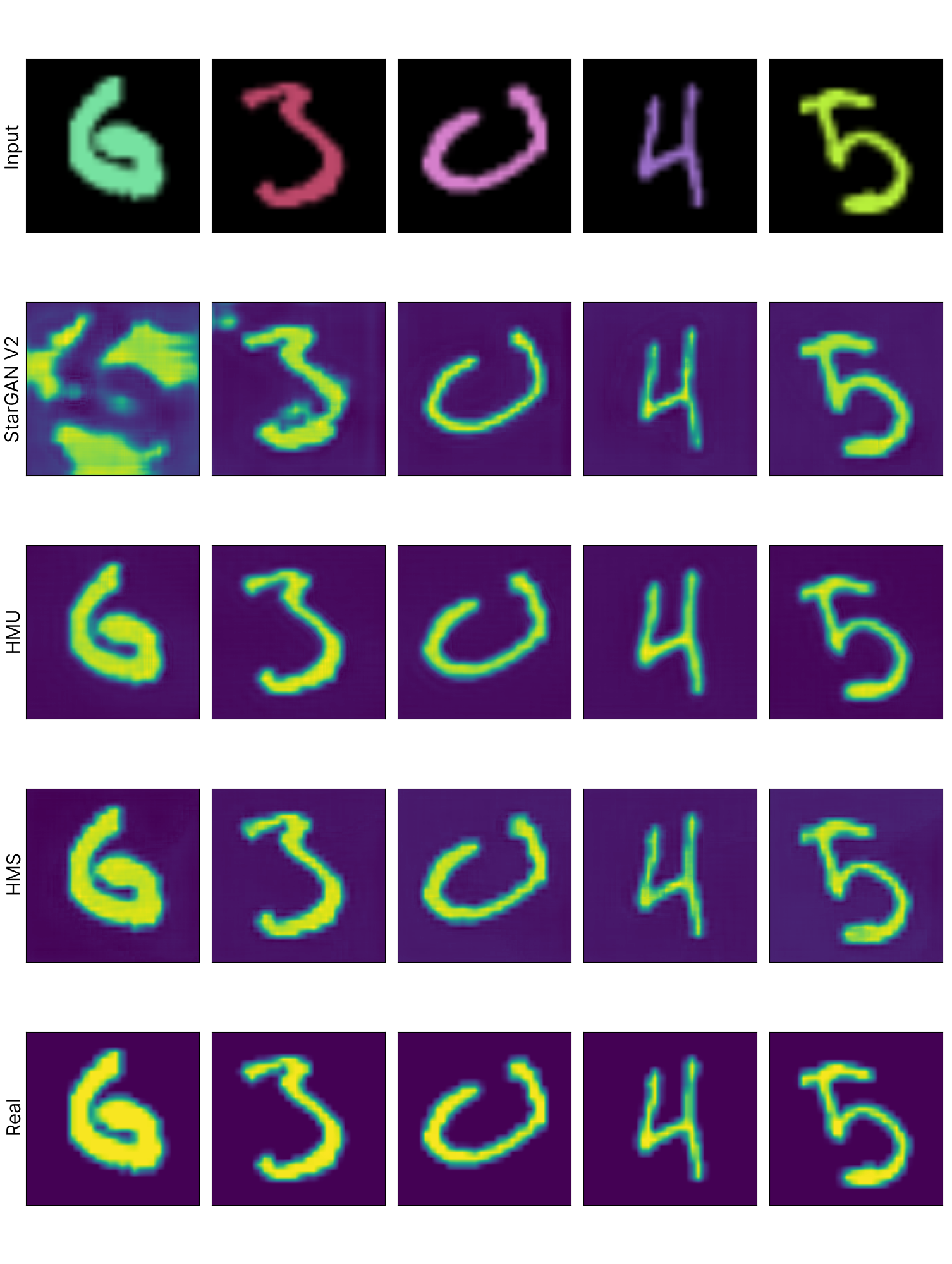}
    \end{subfigure}
    \hfill
    \begin{subfigure}[]{0.4\textwidth}
        \centering
         \includegraphics[width=\textwidth]{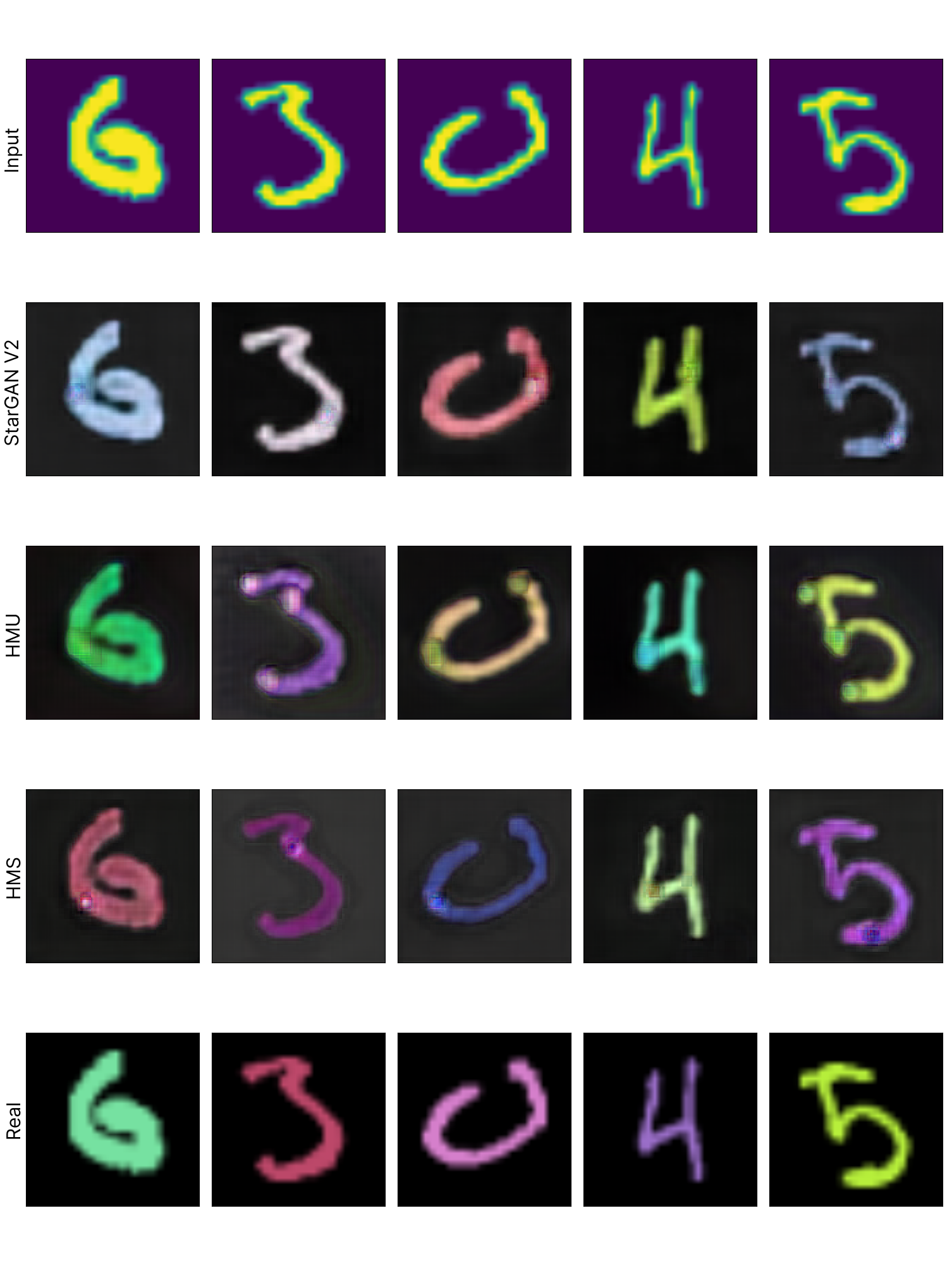}
    \end{subfigure}
    \caption{5 randomly sampled translations on the test set from Colorized-MNIST images to MNIST images (left) and MNIST images to Colorized MNIST images (right).}
    \label{fig:cmnist-to-mnist}
\end{figure}

In general, StarGAN V2, HMU and HMS were all able to successfully learn the mapping between colorized MNIST and MNIST images. Table~\ref{tab:color-recall} shows the results on the Color Recall score for the 3 color channels, an averaged total score, and the number of unique colors that were generated from the test set. Table~\ref{tab:colorized-mnist} summarizes the results on FID, Precision, Recall, Mean Squared Error, and Color-Recall. Figure~\ref{fig:cmnist-to-mnist} shows visual results of image-to-image translation on 5 randomly sampled images from the test set.

In particular, the HMU model was able to outperform the base StarGAN V2 model with a lower MSE while only having a small decrease in color recall. The HMU model was also able to outperform StarGAN V2 on recreating red and blue and was able to capture a wider color range on those 2 channels. 

Both the HMU and HMS models were able to improve on the original StarGAN V2 model by a factor of 10 on generating MNIST images. While the original model had marginally better FID, Precision, and Recall scores, the HMU model was able to achieve a better trade-off (a small loss in diversity in the multi-modal domain with a much larger improvement in mean squared error on the uni-modal domain). With the extra supervised loss term, the HMS model was able to achieve a near-similar performance on the mean squared error to the HMU model, but it also led to a larger loss in diversity. It's possible that by reducing the supervised loss weight, the model would have improved performance. 

To show the application of the Color Recall score, Table~\ref{tab:colorized-mnist} also includes additional results on two popular architectures: CycleGAN \cite{cyclegan}, an unsupervised Generative Adversarial Network for image-to-image translation, and Palette \cite{palette}, a supervised image-to-image translation conditional Diffusion Model. Although it is expected that CycleGAN performs poorly on the Color Recall score (it is known that CycleGAN is not able to produce diverse image distributions), Palette was proposed as a network that is able to produce images with high fidelity and diversity. 

Figure~\ref{fig:color-recall-dist} shows the distributions of the red, green, and blue values that were used to generate the scores. From this, it is simple to see that CycleGAN oversamples high-intensity blue values and Palette oversamples high-intensity green values. This indicates that CycleGAN colorizes MNIST images with different blue tints and Pallete colorizes MNIST images with different green tints. Palette, in particular, oversamples high intensity values for all channels and captures a very small part of the color distributions.

\begin{table}[]
    \centering
    \begin{tabular}{l|c|c|c|c|c}
         Model & Red & Green & Blue & Recall & Count\\\hline
         StarGAN V2 & .6235 & \textbf{.6420} & .6183 & \textbf{.6279} & \textbf{9374}\\
         HMU & \textbf{.7037} & .3843 & \textbf{.7126} & .6002 & 9357\\
         HMS & 0.5774 & 0.6152 & 0.5204 & 0.571 & 8743\\\hline
         CycleGAN \cite{cyclegan} & .5397 & .5590 & .1307 & .4098 & 2474\\
         Palette \cite{palette} & .1324 & .0156 & .2402 & .1294 & 1096\\
    \end{tabular}
    \caption{Color Recall Scores}
    \label{tab:color-recall}
\end{table}

\begin{table}[]
    \centering
    \begin{tabular}{l|c|c|c|c|c|c|c|c}
         Model & \multicolumn{4}{|c}{Colorized MNIST} & \multicolumn{4}{|c}{MNIST} \\\
         & FID & F-8 & F-1/8 & C-Recall & FID & F-8 & F-1/8 & MSE\\\hline
         StarGAN V2 & \textbf{54.00} & \textbf{.5860} & \textbf{.7404} & \textbf{.6279} & \textbf{60.46} & \textbf{.4518} & \textbf{.7967} & .1016\\
         HMU & 60.83 & .4610 & .5888 & .6002 & 62.03 & .2760 & .4434 & \textbf{.0094}\\
         HMS & 89.34 & .3003 & .5482 & .5204 & 81.49 & .3782 & .6571 & .0109\\ 
    \end{tabular}
    \caption{Colorized MNIST $\leftrightarrow$ MNIST Task Results}
    \label{tab:colorized-mnist}
\end{table}

\subsection{ADE20K Results}
\label{sec:ade20k-results}

\begin{figure}
    \centering
    \begin{subfigure}[]{0.4\textwidth}
        \centering
        \includegraphics[width=\textwidth]{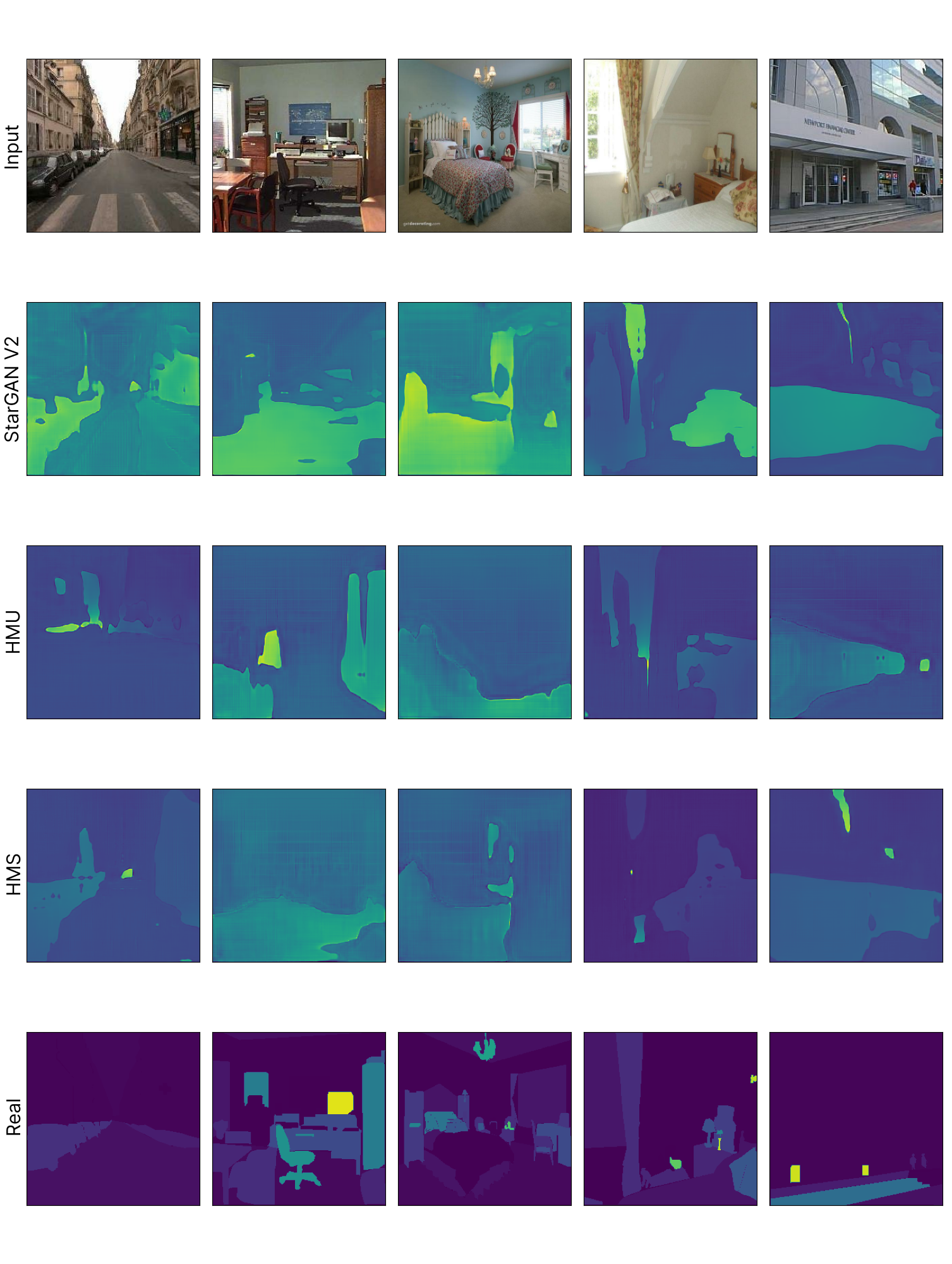}
    \end{subfigure}
    \hfill
    \begin{subfigure}[]{0.4\textwidth}
        \centering
        \includegraphics[width=\textwidth]{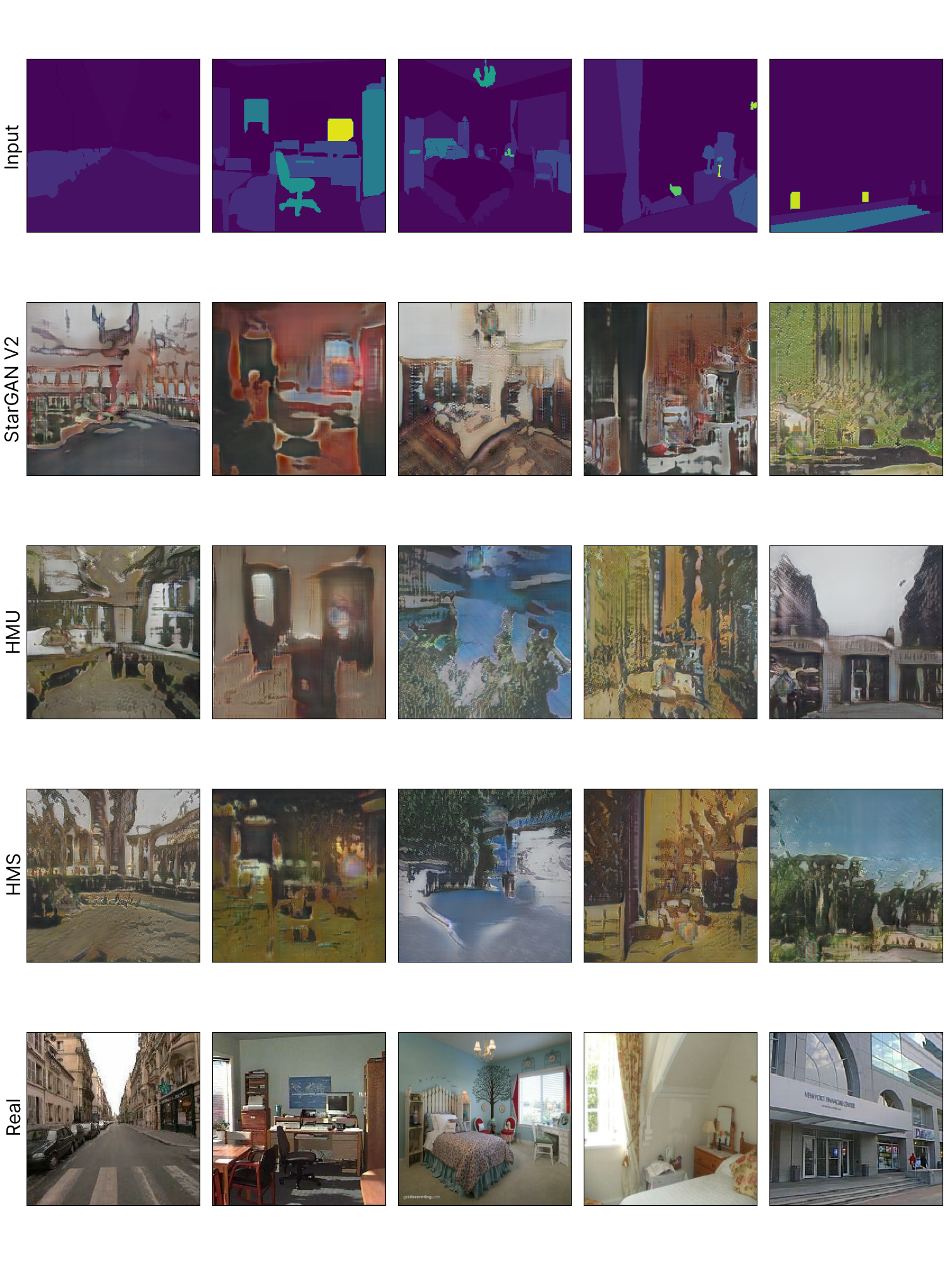}
    \end{subfigure}
    \caption{5 randomly sampled translations on the test set from ADE20K scenes to ADE20K label maps (left) and ADE20K label maps to ADE20K scenes (right).}
    \label{fig:scene-to-label}
\end{figure}

As expected, the ADE20K dataset was significantly more challenging than the Colorized MNIST dataset. Table~\ref{tab:ade20k-results} summarizes the results on FID, Precision, Recall, Mean Squared Error, and Accuracy. Figure~\ref{fig:scene-to-label} shows the visual results of image-to-image translation on 5 randomly sampled images from the test set.

The HMS model was able to outperform the baseline StarGAN V2 model on all metrics except for the F-1/8 score on the generated labels. The HMU model was also able to outperform the original model on all scene generation metrics. Notably, the MSE for the generated label maps decreases with the optimizations in HMU and further decreases with the supervised loss term in HMS. However, in all experiments, the label classification accuracy on the semantic segmentation task is low. This is expected as the model is not optimized for a semantic segmentation task. Label maps are represented as 2D maps with each pixel value corresponding to the class instead of 3D maps with one-hot vectors representing the classes and an L2 loss is used rather than a Categorical Cross Entropy loss.

Even with these additions, the StarGAN V2 generator architecture is not a good model to generate label maps. In Table~\ref{tab:classification}, only the generator of the StarGAN V2 model is trained with an additional classification head that outputs logits for each class and penalized with a Sparse Categorical Cross Entropy loss. Results are included for 2 configurations: one configuration where the weights are initialized with the Xavier Uniform Initializer and one configuration where the weights of the HMS model are used to initialize the model. While using the HMS generator as pre-trained weights performs slightly better than without pretrained weights, the model is only able to achieve around a 7\% accuracy in both configurations. This indicates that the StarGAN V2 generator is a poor model for generating label maps; choosing an alternative architecture that has been designed for a semantic segmentation task may help develop an improved generative network for both the scene generation and label classification tasks. 

\begin{table}[]
    \centering
    \begin{tabular}{l|c|c|c|c|c|c|c|c}
         Model & \multicolumn{3}{|c}{Scenes} & \multicolumn{5}{|c}{Labels} \\\
         & FID & F-8 & F-1/8 & FID & F-8 & F-1/8 & MSE & Accuracy\\\hline
         StarGAN V2 & 159.27 & .1873 & .1214 & 207.45 & .0910 & \textbf{.0778} & .4459 & .0008\\
         HMU & 153.08 & .2417 & .2162 & 221.99 & .0565 & .0325 & .3373 & .0009\\
         HMS & \textbf{146.29} & \textbf{.3095} & \textbf{.2745} & \textbf{193.94} & \textbf{.1022} & .0765 & \textbf{.2787} & \textbf{.0013}\\ 
    \end{tabular}
    \caption{ADE20K Scenes $\leftrightarrow$ ADE20K Labels Task Results}
    \label{tab:ade20k-results}
\end{table}

\begin{table}[]
    \centering
    \begin{tabular}{l|c}
         Pre-trained Weights & Accuracy\\\hline
         None & .0685\\
         HMS & .0724\\
    \end{tabular}
    \caption{Classification model accuracy on ADE20K dataset.}
    \label{tab:classification}
\end{table}

\subsection{Limitations and Future Work}

The approach outlined in Section~\ref{sec:opt} is limited to models that disentangle style and content space. This approach could extend to other methods, such as \cite{oasis}, that use noise vectors to introduce diversity by setting the noise to 0 for uni-modal domains, but this has not been evaluated in this paper. The Color Recall results on Pallete, however, may indicate that disentangling content and style spaces is necessary to produce diverse image distributions. 

All models did not perform well on the ADE20k dataset; however, the goal of this paper was not to introduce a new state-of-the-art technique for generating ADE20k images, but instead to propose a general approach to improving any generative model on many-to-one datasets. 

Future work should expand this work by using better base model architectures (instead of StarGAN V2) to achieve state-of-the-art performance on more complex benchmark datasets \cite{ade20k1, ade20k2, nyu-depth, synthia, imagenet}. Although the focus of this paper was not on pre-training, a promising application of many-to-one generative models could be to pre-train classification (i.e. semantic segmentation) or regression (i.e. depth estimation) models on tasks with limited labeled data. In many-to-one translation tasks, it can be less expensive to collect data for one domain than in another. Using the unsupervised HMU approach, it would be possible to pretrain with imbalanced datasets (for example, many real images and very few LiDAR images for depth estimation) and then fine-tune the weights for these tasks.

\section{Conclusion}

In this work, we have introduced a new framework to optimize generative image-to-image translation models for many-to-one tasks. We have shown that by applying this framework to StarGAN V2, we are able to acheive an improved joint performance on both the multi-modal and uni-modal domains. We have also introduced a new benchmark dataset and associated metrics that can be used to provide interpretable results on the ability of a model to perform many-to-one tasks. 
\medskip

{\small
\bibliographystyle{ieee_fullname}
\bibliography{refs}
}


\end{document}